\documentclass{article}
\pdfoutput=1

\usepackage{arxiv}

\usepackage[utf8]{inputenc} 
\usepackage{hyperref}       
\usepackage{url}            
\usepackage{booktabs}       
\usepackage{amsfonts}       
\usepackage{nicefrac}       
\usepackage{cleveref}       
\usepackage{graphicx}
\usepackage{natbib}
\usepackage{doi}
\usepackage{authblk}

\title{LARP: Language-Agent Role Play for Open-World Games} 

\date{}

\newif\ifuniqueAffiliation
\uniqueAffiliationtrue

\ifuniqueAffiliation 

\author[1]{Ming Yan \footnote[1] \footnote[2]}
\author[1]{Ruihao Li \footnote[1]}
\author[1]{Hao Zhang}
\author[1]{Hao Wang}
\author[1]{Zhilan Yang}
\author[1]{Ji Yan}

\affil[1]{MiAO}

\renewcommand{\headeright}{}
\renewcommand{\undertitle}{}
\renewcommand{\shorttitle}{}

\hypersetup{
pdftitle={LARP: Language-Agent Role Play for Open-World Games},
pdfsubject={Language Agent for Role Play},
pdfauthor={Ming Yan},
pdfkeywords={},
}

\begin{document}
\maketitle

\renewcommand{\thefootnote}{\fnsymbol{footnote}}
\footnotetext[1]{Equal Contribution}
\footnotetext[2]{Correspondence to: mingyan@miao.company}

\begin{abstract}
    Language agents have shown impressive problem-solving skills within defined settings and brief timelines. Yet, with the ever-evolving complexities of open-world simulations, there's a pressing need for agents that can flexibly adapt to complex environments and consistently maintain a long-term memory to ensure coherent actions. To bridge the gap between language agents and open-world games, we introduce Language Agent for Role-Playing (LARP), which includes a cognitive architecture that encompasses memory processing and a decision-making assistant, an environment interaction module with a feedback-driven learnable action space, and a postprocessing method that promotes the alignment of various personalities. The LARP framework refines interactions between users and agents, predefined with unique backgrounds and personalities, ultimately enhancing the gaming experience in open-world contexts. Furthermore, it highlights the diverse uses of language models in a range of areas such as entertainment, education, and various simulation scenarios. The project page is released at \url{https://miao-ai-lab.github.io/LARP/}.
\end{abstract}


\section{Introduction}
\label{sec:introduction}
Large Language Models (LLMs) are machine learning models capable of performing a variety of Natural Language Processing (NLP) tasks such as generating text, translating text from one language to another, and answering questions conversationally. The term \textit{large} refers to the vast amount of parameters the language model can update during its learning process. With the development of pre-training generative model techniques and the construction of massive and comprehensive datasets, some top-performing large language models have up to hundreds of billions of parameters \citep{touvron2023llama, radford2018improving, radford2019language, brown2020language, ouyang2022training, openai2023gpt4}. Furthermore, owing to the advancements in large language models, AI entities have become a hot topic in recent years. These artificial intelligence entities, often referred to as \textit{agents} \citep{russell2010artificial, wooldridge1995intelligent}, are the fundamental components of large-scale artificial intelligence systems. Typically, in the realm of Artificial General Intelligence, an agent is an artificial entity that can perceive its surrounding environment through sensors, make decisions, and respond via actuators. With the development of large language models and agents, there is a new trend of combining them into a single entity called language agents. These language agents are designed by integrating large language models and agent design \citep{wang2023survey, xi2023rise, sumers2023cognitive}. \\

As a strongly related industry to computers, gaming has become increasingly intertwined with the development of general-purpose language agents. The application of LLMs and agents has grown more widespread. In related studies, there is considerable literature on the application of language agents in text-based games \citep{dambekodi2020playing, singh2021pre, yao2021reading, urbanek2019learning} and adversarial games \citep{openai2019dota, arulkumaran2019alphastar}. Concurrently, with the enhanced capabilities of LLMs, open-world games have emerged as the frontier for language agent applications. This is due to the unique and challenging scenarios present in open-world games, which provides a fertile ground for general-purpose language agents. Open-world games present a rich, dynamic, and engaging environment, encompassing complex missions and storylines. They require the use of agents to equip non-player characters with diversified behaviors. Although numerous studies have proposed architectures for general language agents that could be applied in open-world games like Minecraft \citep{lin2023agentsims, park2023generative}, a gap still exists between the general-purpose agents and the overall requirements in an open-world gaming context. General-purpose language agents are created to address a variety of issues in realistic environments. Their primary requisites are universality and the simulation of human behavior. These agents can adapt to various environments and tasks, without being restricted to fixed roles. However, these general-purpose language agents face significant challenges in practical open-world environments. These challenges include but are not limited to interpreting complex environments, memorizing long-term events, generating expressions that cohere with character and environmental settings, and continuously learning from interactions with the environment. \\

Hence, in this work, we propose a game-oriented Role-Playing Agent framework, Language Agent for Role Play (LARP) toward open-world games. LARP focuses on blending the open-world games with language agents, utilizing a modular approach for memory processing, decision-making, and continuous learning from interactions. In the agent's internal depiction, we designed a complex cognitive architecture based on cognitive psychology, equipping agents under the LARP framework with high playability and uniquity. Aiming to yield more realistic role-playing experience, we regularized agents using the data and context of the open-world gaming environment, prior set personalities, knowledge, rules, memory, and post constraints, which can be seen as a specific case within the general-purpose language agents. As for the general agent architecture, it typically requires a large-scale language model. However, our architecture incorporates a cluster of smaller language models, each fine-tuned for different domains, to handle various tasks separately. This design contributes new experiences and perspectives to developing language agents for open-world role-playing games. \\

\begin{figure}
	\centering
    \includegraphics[width=1.0\textwidth]{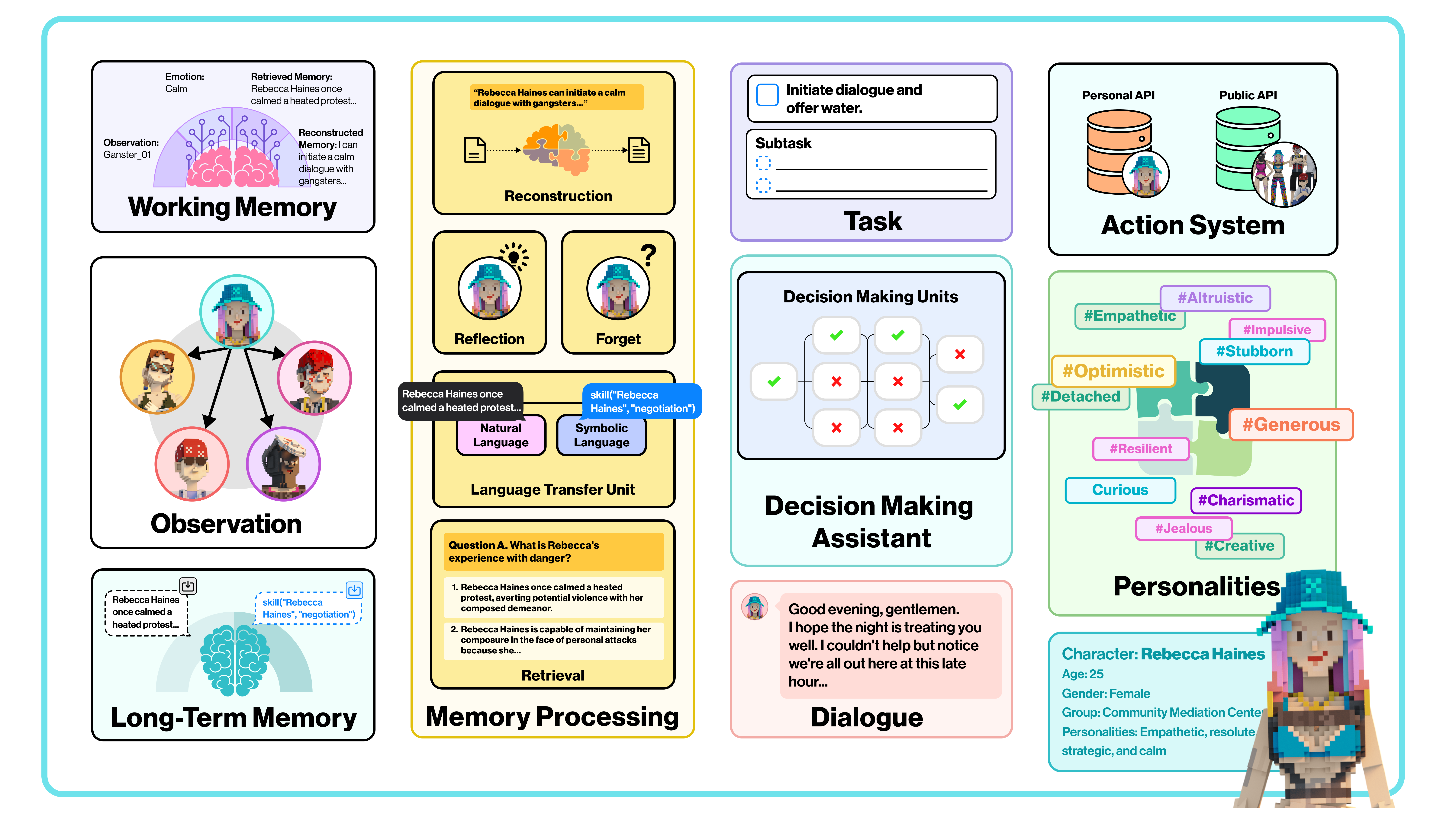}
	\caption{Cognitive Architecture of LARP Overview.}
	\label{fig:fig_overflow}
\end{figure}

The rest of this work is organized as follows: Section \ref{sec:related} discusses the related works on agent framework and agent components. In Section \ref{sec:cognitive}, we present the cognitive architecture part of LARP. In Section \ref{sec:environment}, we highlight the environmental interaction module, and in Section \ref{sec:diversity}, we introduce agents' alignment of diversified personalities. A discussion is presented in Section \ref{sec:discussions}, followed by a conclusion of the entire paper in Section \ref{sec:conclusion}. \\
\section{Related Work}
\label{sec:related}
The development and advancements in LLMs have brought opportunities and potential to various fields. These fields include but are not limited to, role-playing, gaming, and tool-using Agents, amongst others. At the same time, the definition of language agents is being continuously updated. We will share some related works on agent architecture in this section. \\

\subsection{Agent Framework}
Firstly, we will introduce some works related to language agents' role-playing and simulation, which are aimed at enhancing the LLM's ability in role-playing and highlighting distinct features. These works also aim to boost the interaction capability between the agents and users, making the agents appear more \textit{self-conscious} \citep{wang2023rolellm, shao2023character, shanahan2023role, li2023chatharuhi}. Other studies focus on role-playing and interaction among multiple agents, which include scenarios such as collaboration to complete tasks \citep{li2023camel, chen2023gamegpt, qian2023communicative, li2023metaagents, wu2023autogen}, simulating daily activities \citep{lin2023agentsims, park2023generative, wang2023recagent, liu2023training}, and promoting progress in debates \citep{liang2023encouraging, du2023improving, chan2023chateval}. \\

In addition to this, language agents are also applied in open-world environments. There are not only instances of application in text-based games \citep{urbanek2019learning, cote2019textworld, hausknecht2020interactive}but also exploration tasks in open-world environments such as Minecraft \citep{wang2023describe, wang2023voyager, zhu2023ghost}. \\

\subsection{Agent Component}
In this section, we will introduce some related works on the design of language agent components. An agent system is usually divided into three parts: memory, planning, and action (tool use) \citep{weng2023prompt}. Memory system serves as a repository for facts, reflections, etc., entailing abilities for storage and retrieval. Hence, efforts in memory primarily pertain to input/output functions, including memory compression \citep{hu2023chatdb}, storage, and retrieval \citep{park2023generative, zhong2023memorybank, huang2023memory}.  \\

The planning component is responsible for the decision-making aspect related to the agent's behavior and language. The capability of agents largely depends on this part. Abilities for planning \citep{yao2023tree, liu2023llm+, yao2022react, shinn2023reflexion, liu2023chain, wang2023plan} and reasoning \citep{wei2022chain, madaan2023self} are realized in this component, and associated works typically revolve around these two abilities. \\

The last component is tool usage and actions, which signifies an augmentation in the capabilities of the intelligent entities, aiding them in conducting more complex and difficult tasks. Works in this section include tool-using \citep{nakano2021webgpt} and learning new actions \citep{schick2023toolformer}. \\
\section{Cognitive Architecture}
\label{sec:cognitive}
\begin{figure}
	\centering
    \includegraphics[width=1.0\textwidth]{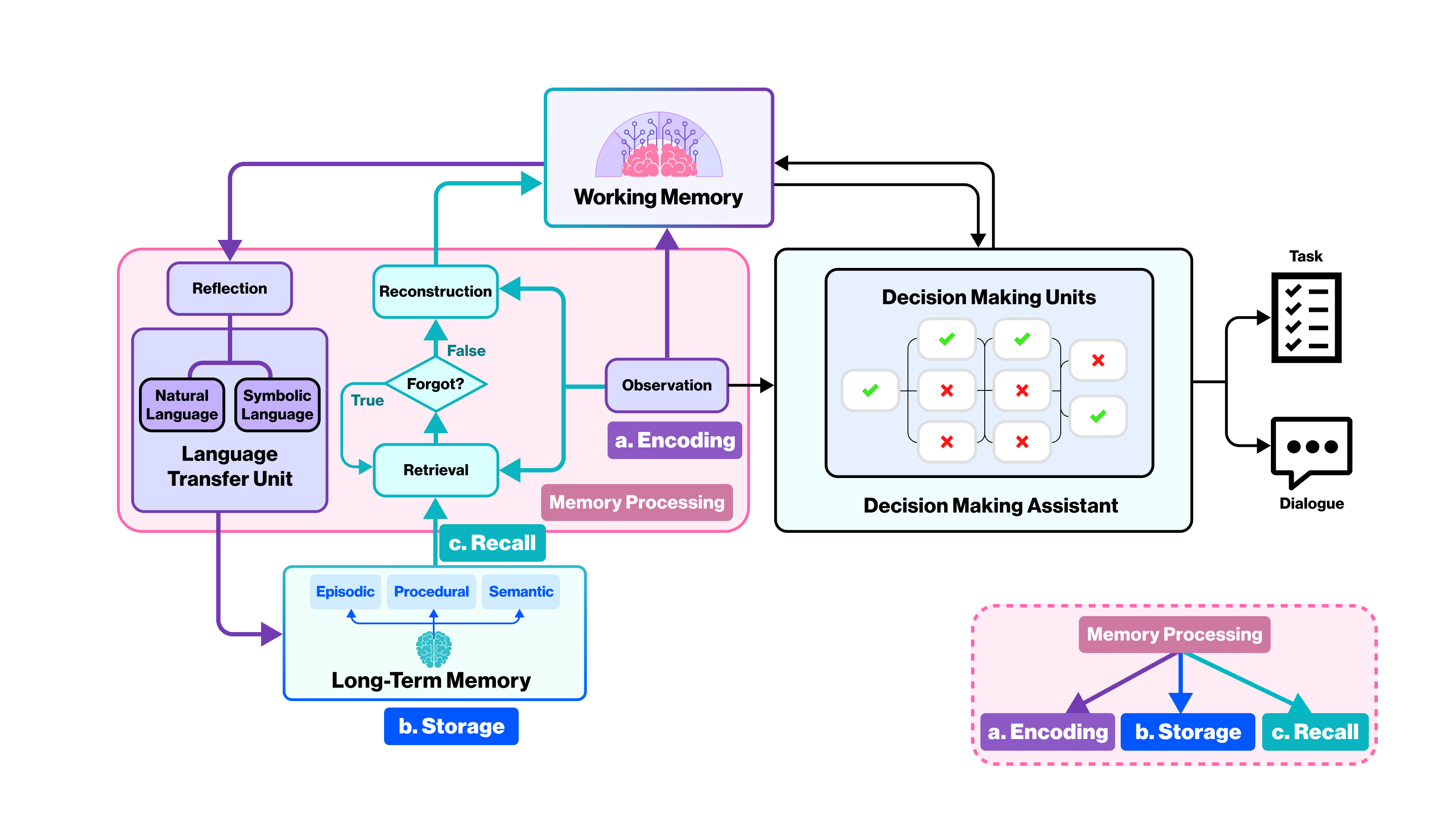}
	\caption{Cognitive Workflow of LARP. This represents a cycle: Information from long-term memory and observation is processed in the memory processing module and transmitted to the working memory module. The information in the working memory module, together with the observed information, is inputted into the decision-making assistant, which finally generates a decision or dialogue. Memory processing has three main stages: encoding, storage, and recall. Encoding is the process of transforming information into a form that can be stored in memory. Storage is the process of maintaining information in memory. Recall is the process of retrieving information from memory. }
	\label{fig:fig1}
\end{figure}

    Cognitive architecture is a fundamental component of role-playing language agents in open-world games. It provides a logical framework and enables self-recognition of agents. The cognitive architecture is shown in figure \ref{fig:fig1}. It comprises four major modules: long-term memory, working memory, memory processing, and decision-making. The long-term memory module serves as the primary warehouse containing memories with substantial storage capacity. Working memory acts as a temporary cache with limited space for memory. The memory processing module is the most important unit of the cognitive architecture. The decision-making module then derives agents' subsequent actions based on the retrieved information. \\

\subsection{Long-Term Memory}
    In cognitive science, long-term memory (LTM) is comprised of two types of memory: declarative memory and procedural memory. Declarative memory is further divided into semantic memory and episodic memory \citep{laird2019soar, tulving1972episodic}. Semantic memory refers to general knowledge memories acquired through conceptual and factual knowledge about the world. In the context of the open-world game, it can be considered the part that encapsulates the game rules and memories consistent with the relevant worldview. We divided semantic memory into two parts in the system. One is implemented with an external database, as its content is not frequently changed. Simultaneously, some semantic memories are stored in the long-term memory module in symbolic language. \\

    Episodic memory refers to the memory of specific events that an individual experiences. These can be memories related to other players or Agents. In our memory system, we adopted a vector database in the long-term memory module for storing and retrieving these memories. Associated decay parameters are introduced as memories can be forgotten, with relevance scores decreasing over time. When reasoning with the LLM, such memory contents can be easily retrieved through vector queries. \\

    Procedural memory refers to actions or skills that can be performed without conscious thought\citep{roediger1990implicit}, such as swimming, cycling, etc. These skills, with action properties, are represented as APIs in action space in our system. The action space is divided into public APIs and personal APIs. The personal APIs could be extended through learning \citep{sumers2023cognitive} which is mentioned in the section \ref{sec:environment}. \\

    In the long-term memory module, we store all perceived memories in the semantic and episodic memory zones respectively. We propose a method named \textbf{Question-based Query}, which generates self-ask questions as queries that can be leveraged in search by vector similarity and predicate logic. This method facilitates the retrieval of semantic and episodic memories in the recall module, thereby improving the overall efficiency of memory utilization. \\

\subsection{Working Memory}
    Working memory mainly holds the observational information and retrieved long-term memories needed for performing complex cognitive tasks (such as reasoning and learning) and interactive tasks \citep{baddeley2003working, miller2017plans}. These pieces of information are typically obtained through agents' observation as natural language data provided by the game side. Short-term memory, as its name suggests, represents a stage of memory that retains information for brief periods of time, generally only lasting a few seconds to a minute \citep{atkinson1968human}. To humans, the average capacity for retaining items in short-term memory is about 7±2, with a retention duration of roughly 20 to 30 seconds \citep{miller1956magical}. In this work, the two concepts are implemented as the same module, collectively referred to as working memory. In our architecture, it exists as a data cache from which information is extracted and dropped into the prompt's context. Its extraction process is further explained in more detail in the memory processing and decision-making sections.  \\

\subsection{Memory Processing}
    The memory processing module primarily handles the processing of memories that have been stored and are about to be stored. The three main stages of memory are encoding, storage, and recall \citep{melton1963implications}. Specifically, perceived input information is encoded and transformed into content in long-term memory, enabling it to be recalled in the space of long-term memory. In LARP, we simulate this procedure by processing all structured observational information provided in the game, combining it with retrieved content, and storing it in the working memory. This information serves as an input for a series of logic processing units in the decision-making module, continuously updating the content in the working memory. Once the length of working memory reaches a certain threshold, reflection is triggered, during which ineffective memories are filtered out, and the processed natural language memories and symbolic language memories are separately stored as episodic memory and semantic memory. \\

    The core of memory encoding is the language transformation system. By aligning language models and probabilistic models, natural language is converted into a probabilistic programming language (PPL) \citep{wong2023word} and logic programming language. PPL primarily handles probabilistic reasoning, while logic programming language pertains mainly to fact reasoning. Moreover, memory encoding should also be influenced by the consistency of prior knowledge, meaning that past knowledge will affect the current understanding \citep{bartlett1995remembering}. The storage of memory has already been elaborated in the long-term memory section. \\

\begin{figure}
	\centering
    \includegraphics[width=1.0\textwidth]{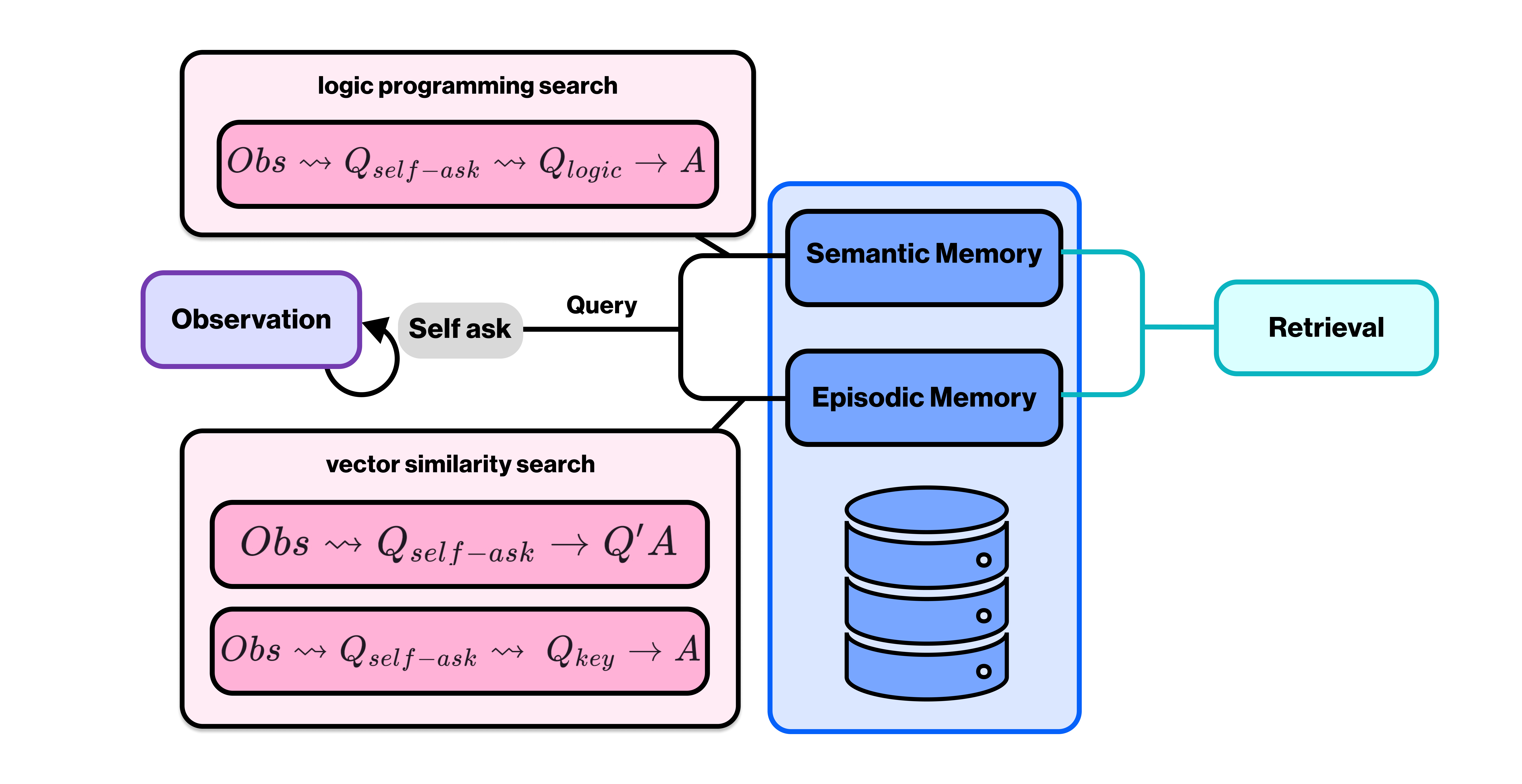}
	\caption{Detail control flow of recall psychological process. First conduct self-asking about the observation to get self-ask questions. Using the self-ask questions as queries, different methods of retrieval are undertaken. 1. Generate predicate logic statements in logic programming language and probabilistic programming language based on queries. 2. Conducting a vector similarity search after extracting keywords from the queries. 3. Searching for question-answer pairs based on sentence similarity between queries and the questions of question-answer pairs. $Q_{self-ask}$ means the self-ask questions which were used as queries, $Q_{logic}$ stands for predicate logic query statements, $Q_{key}$ is the extracted keywords, ${Q'A} $ stands for the question-answer pairs. }
	\label{fig:fig2}
\end{figure}

    To humans, recall refers to the psychological process of retrieving information from the past. While in our architecture, it is the process of retrieving information from long-term memory. It first involves compound retrieval from long-term memory, including search by vector similarity and predicate logic. First, we employed self-ask strategies to form the queries, prompting LLM to raise questions regarding agents' observations, personalities, and experiences. After obtaining the queries, we adopted 3 methods to perform the retrieval. For logic programming search, LLM generates a query in a logic programming language that answers the self-ask questions based on available rules and facts. For similarity search, two methods were available. One method is using self-ask questions as queries for vector similarity search, matching with question-answer pairs in the vector database of episodic memory. The other method is using keywords extracted from the self-ask questions to match with the natural language memories in the same database. This process will be repeated until a final answer is obtained, and this can also be considered semantic retrieval \citep{press2022measuring}. Figure \ref{fig:fig2} shows the detailed control flow. \\

    Based on the recall capabilities, our architecture adopts CoT \citep{wei2022chain} to reason about the retrieved content and observed information and perform memory reconstruction, ie., using prior knowledge to influence observed facts to some extent \citep{loftus1974reconstruction} though the reconstructed memories might be distorted. In addition, we also simulated the process of human forgetting in the recall workflow. When the retrieval system operates, we introduce a decay parameter $\sigma$ represented by Wickelgren's power law to mark the forgetting probability of this memory \citep{wixted2007wickelgren}. The calculation formula is as follows: \\

    \begin{equation}
        \sigma = \alpha \lambda N(1 + \beta t)^{-\psi}
    \end{equation}

    Here, $\lambda$ represents the importance level, given by a scoring model. $N$ stands for the number of retrievals of this memory, and $t$ is the elapsed time after the last retrieval. $\psi$ is the rate of forgetting for each character. $\alpha$ and $\beta$ are the scaling parameters for importance and time, respectively. Through multiple rounds of memory reconstruction and forgetting processes, our cognitive architecture can ultimately simulate instances of memory distortion. \\

\subsection{Decision Making}
    The decision-making module produces final decisions under the joint effect of observation and working memory. The core section of the decision-making module is an ordered cluster of programmable units. Each unit will process the content in working memory and context, updating the results to the working memory in real time. These units can be simple information processing units, such as those conducting affective computing, or complex units equipped with specifically fine-tuned LLM models, like intent analysis and output formatting. These units are infinitely scalable and can handle all types of memory processing tasks. As each unit communicates with working memory, it updates the working memory in real-time, allowing the agents to react timely when observation changes during the process. The execution order of these units would be determined by a language model assistant. The final output of the decision-making module could be tasks or dialogue contents for the Non-Player Characters (NPCs).\\

\section{Environment Interaction}
\label{sec:environment}
For role-playing language agents in open-world games, generating tasks based on current observations through the cognitive architecture only accomplishes the objective within agents. However, in open-world games with free actions and rich gaming content, agents need to interact with the game environment by connecting the internal and the external. There are various works in employing language agents to interact with open-world game environments \citep{wang2023describe, zhu2023ghost, yang2023octopus, wang2023voyager}. For instance, Voyager uses the concept of an \textit{automatic curriculum}, obtaining objectives by feeding GPT4 the contents and states of environmental observations. Then, GPT4 is prompted to generate the functioning code to reach the objectives. The paper also presents the \textit{skill library} method, which embeds the description of the generated code as a key and the code as a value, achieving high extensibility of incorporating new skills by adding key-value pairs. Octopus utilizes the Visual Language Model (VLM) to acquire observations. However, this approach can lead to high dimensions of data feature distribution, resulting in poor controllability. Moreover, the operational cost of the VLM is high, and it's challenging to collect the data set's prior knowledge within the game. \\

\begin{figure}
	\centering
    \includegraphics[width=1.0\textwidth]{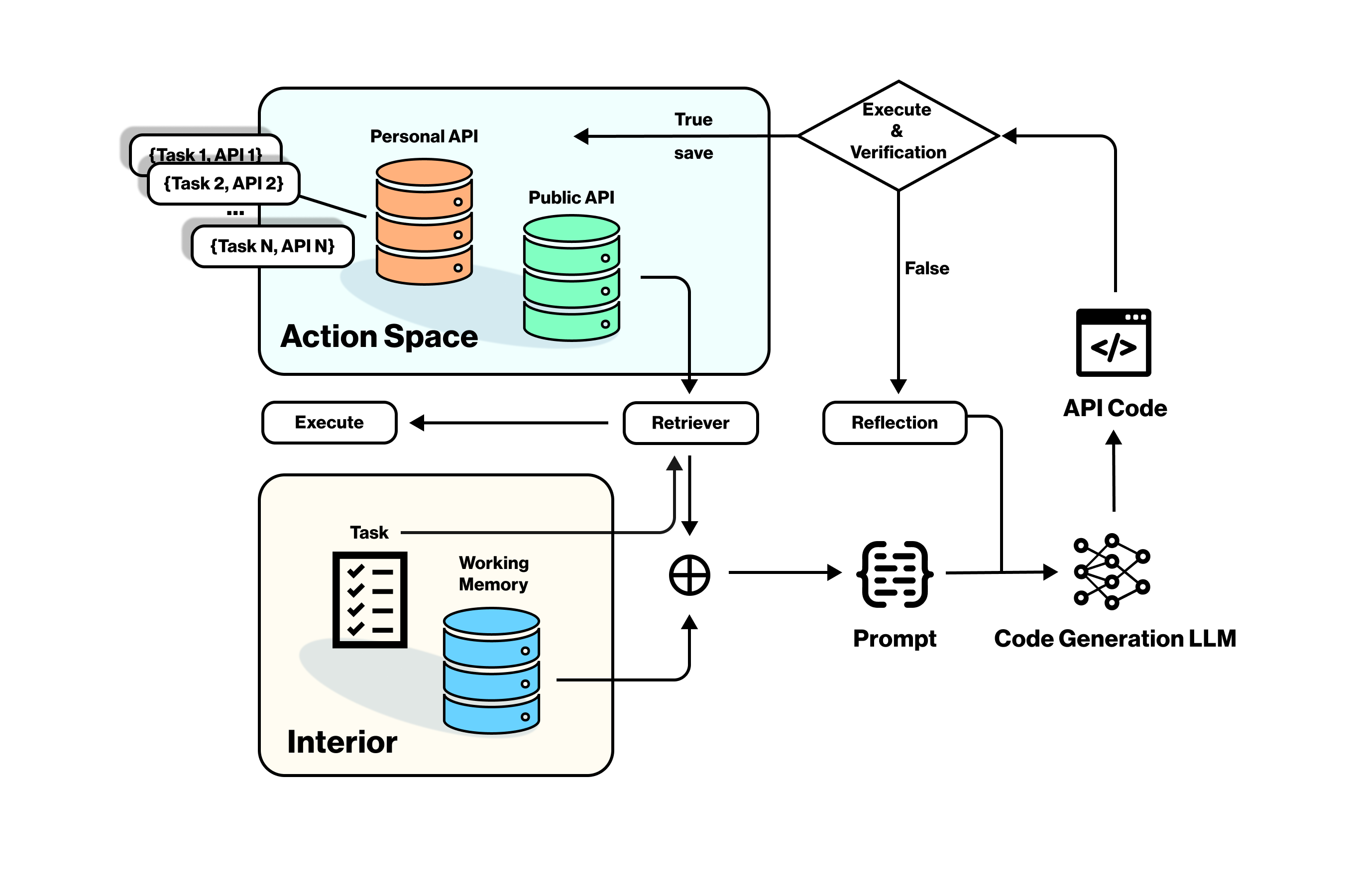}
	\caption{Environment Interaction.}
	\label{fig:fig3}
\end{figure}

Figure \ref{fig:fig3} shows the fundamental interaction procedural. Interior refers to the working memory and the tasks that need to be executed based on the current situation, generated by the observation and cognitive architecture. The Action Space is the agent's executable action APIs in the game world, encompassing both public and personal APIs. The personal API library stores tasks-API pairs, while the public APIs are basic actions. A personal API can be a series of basic actions, facilitating quick decision-making and reuse of APIs. \\
 
Once we have generated the corresponding plans in the decision-making module, we initially attempt to break down the overall task goal into several subtask goals. These subtask goals present as strictly ordered sequence-sensitive arrangements. For each task goal or subtask goal, the whole system will integrate it with the working memory. Then, it will use a retriever to search separately in the personal API library and public API library. If the action corresponding to the task already exists in the personal API library, the action is instantly performed. Otherwise, the system completes the corresponding prompt with the entire action space and interior content to generate the structured code using a fine-tuned LLM. Upon the successful execution and verification of the generated code blocks, they are stored as a new interface in the personal API library in the form of (Task, API) for future use. If verification fails, the reflection unit is activated to generate new code blocks \citep{shinn2023reflexion}. \\

Simultaneously, we also collect the paired prompt and generated code as a training set for fine-tuning code generation LLM \citep{patil2023gorilla}. After the successful execution and verification, the results are fed back through RLHF to enhance the model's capabilities. \\

\section{Personalities}
\label{sec:diversity}
The role of distinct personalities is vital in enhancing the cognitive capabilities of language agents in role-playing. Aligning variable personalities permits language models to better comprehend different perspectives and portray various cultures and social groups. Language models acting out different roles in complex scenarios must deeply understand, respond, and express in their unique ways. This necessitates the models to possess human-like thought processes with a wide range of personalities. Understanding the generation of diverse language expressions, handling multicultural content, and exhibiting varied thoughts, perspectives, emotions, and attitudes - all demand the model to accommodate these distinct personalities. Therefore, this section will delve into its implementation.

LARP adopts the strategy of simulating a cluster of models fine-tuned with various alignments to tackle the agents' diversified viewpoints. These models might be applied in different modules. During the training phase, we pre-trained several base models of different scales. Our pretraining datasets contain perspectives of different cultures and groups. After pre-training, these base models undergo supervised fine-tuning (SFT) on a series of instruction datasets of persona and character to enhance instruction-following and role-playing capabilities \citep{chen2023maybe, dong2023abilities}. This instruction dataset was established through data distillation based on question-answer pairs generated by SOTA models. Then, the dataset was optimized via assessment, modification, and adjustment based on human feedback. \\

It is possible to create multiple datasets and fine-tune LoRAs (Low-Rank Adaption) for capabilities such as reflection, code generation, and intent analysis. These LoRAs can be dynamically integrated with base models of different scales, creating a model cluster with diverse capabilities and personalities. These capabilities cover tasks such as language style, emotion, action generation, reflection, memory reconstruction, and more. \\

However, one of the main challenges in fine-tuning language models to construct different LoRAs for role-playing is acquiring quality data. Successful fine-tuning requires custom datasets of high quality, which need to be carefully constructed to capture various aspects of characters, including their language style, behavior style, personality traits, idioms, backstories, and more. The construction of datasets requires extensive literary creativity, script compilation, and character research to ensure that the generated language not only fits the character's persona and features but also interacts with the user in an appropriate manner \citep{wang2023rolellm}. \\

To enrich the diversity of the agent, we set up several post-processing modules, including the action verification module and the conflict identification module. The action verification module is part of the environment interaction module, which checks whether the generated actions can be correctly executed in the game. Conversely, within the cognitive architecture, the conflict identification module checks whether the decisions and conversations encompass conflicts with character relationships, personalities, and game worldview. When such conflicts are detected, the module will undertake actions like rejecting the result or rewriting it to prevent the agent from going out of character. \\

\section{Discussions}
\label{sec:discussions}

\subsection{Multi-Agent Cooperation and Agent Socialization}

A single Language Agent role-playing under the framework proposed in this work is insufficient to solve the issue of creating rich content in open-world games. To bring each character supported by an Agent to life, a robust social network needs to be established. One possible approach is to build suitable sociological mechanisms and behaviors atop the large language model-driven agents to ensure that NPCs can still maintain their rationality and logic after extensive role-playing reasoning. \\

\subsection{Confidence of Model Clusters vs. Evaluation and Feedback System}

Combining language models and cognitive science makes language agents align more closely with genuine human cognition. This method effectively mitigates the problem of a single large model's inability to enhance role-playing outcomes due to insufficient data. Simultaneously, since the cognitive system only consists of domain tasks, fine-tuned small-scale models can achieve satisfactory performance. It saves costs compared to fine-tuning large models. However, due to the randomness of language model output results, it's unpredictable how the cumulative distortion of the results produced by each task affects the distortion of the entire cognitive architecture. It's hard to say such a distorted agent could be called a human-believable agent. Therefore, a corresponding evaluation and a measurement framework are needed to impose constraints and convergence on the distortion of the cognitive system. Establishing a measurement and feedback mechanism for the entire system to measure the logical deviation of each logic unit can optimize system robustness and minimize the impact of single-system distortion on the overall system. \\

\section{Conclusion}
\label{sec:conclusion}
In this study, we present a language agent framework-oriented open-world games and elaborate on this framework from three aspects: cognitive architecture, environmental interaction, and alignment with diverse value perspectives. Addressing cognitive architecture, we employ more intricate techniques from cognitive science to enable the agent to make more reasonable decisions, alongside implementing post-processing constraints to prevent undue freedom in the agent, thereby bringing them closer to real human behavior in role-playing contexts. We envisage that our work harbors tremendous potential within the open-world games, breathing new life into this traditional domain, and ultimately catering the experience akin to that of 'Westworld'. \\

\bibliographystyle{unsrtnat}
\bibliography{references}  

\end{document}